%% file: body.tex
\documentclass[]{ngsm2024} 
\usepackage{graphicx}
\usepackage{amsmath}
\usepackage{color}
\usepackage{lipsum}
\usepackage{wrapfig}

\vspace{-150mm}
\title[Delay Embedding Theory of Neural Sequence Models]{Delay Embedding Theory of Neural Sequence Models }

\optauthor{%
\Name{Mitchell Ostrow} \Email{ostrow@mit.edu}\\
\Name{Adam Eisen}
\Email{eisenaj@mit.edu}\\
\Name{Ila Fiete} \Email{fiete@mit.edu}\\
\addr Department of Brain and Cognitive Sciences, Massachusetts Institute of Technology}

\begin{document}
\vspace{-800mm}

\maketitle
\vspace{-10mm}
\begin{abstract}

To generate coherent responses, language models infer unobserved meaning from their input text sequence. One potential explanation for this capability arises from theories of delay embeddings in dynamical systems, which prove that unobserved variables can be recovered from the history of only a handful of observed variables. To test whether language models are effectively constructing delay embeddings, we measure the capacities of sequence models to reconstruct unobserved dynamics. We trained 1-layer transformer decoders and state-space sequence models on next-step prediction from noisy, partially-observed time series data. We found that each sequence layer can learn a viable embedding of the underlying system. However, state-space models have a stronger inductive bias than transformers--in particular, they more effectively reconstruct unobserved information at initialization, leading to more parameter-efficient models and lower error on dynamics tasks. Our work thus forges a novel connection between dynamical systems and deep learning sequence models via delay embedding theory. 



\end{abstract}

\input{introduction}

\input{methods}
\input{results}

\input{discussion}

\bibliography{bib}
\newpage
\clearpage
\appendix

\input{appendix}

\end{document}

%% file: introduction.tex
\section{Introduction}

Neural sequence models, specifically transformers and state-space models (SSMs) trained on next-token prediction, have made extraordinary strides in natural language processing \cite{vaswani2017attention,gu2021S4, orvieto2023LRU, gu2023mamba, lin2021transformer-survey, lmu,hippo,s4d}. More generally, these models operate over ordered sequences of data, and thus have the potential to be learners of any temporal prediction problem. Yet, transformers have been noted to underperform in continuous time-series prediction \cite{zeng2023forecasting}, an issue that several transformer architecture variants have sought to rectify \cite{wu2021autoformer, zhou2021informer, nie2022patchtst}. Certain SSMs outperform transformer variants on benchmark time-series prediction tasks \cite{gu2021S4}. Furthermore, it is unclear why transformers underperform other models in time-series forecasting.

In this work, we present mechanistic insights into the performance of transformers and SSMs on time-series prediction tasks of well-characterized dynamical systems. We examine their learned representations and quantify their alignment to the dynamical structure of the underlying system. Our results connect neural networks to the theory of delay embeddings in dynamical systems, thereby shedding light on the inductive biases and capabilities of each architecture for time-series prediction. 

\subsection{Delay Embedding Theory}

Delay embedding is a well-known method for reconstructing and characterizing the geometry of chaotic dynamical systems, when the system is only partially observed. Delay embedding a time-series involves stacking time-delayed copies of the observed data into a vector. The most famous theory in the field, Takens' Delay Embedding Theorem, proved that with sufficiently many delays, a delay embedding of a single variable in a multi-variable dynamical system is diffeomorphic to the original dynamics \cite{takens1981}. However, this theorem lacks two important components for practical use: how to pick the optimal delay embedding parameters, and an understanding of the role of observational noise \cite{casdagli1991reconstruction}. While Takens provided a prescription for a minimal number of delays to reconstruct the attractor, this is not necessarily the optimal delay embedding. Better embeddings integrate enough information to be robust to observational noise, but not too much so that they are overly distorted, thereby hampering downstream prediction. We demonstrate these tradeoffs in Fig.\ref{fig1}. Thus, when creating a delay embedding, one must be selective about which components of the history are used. Methods to pick appropriate delays include autocorrelation analysis, mutual information, persistent homology, and more \cite{packard1980geometry,gibson1992analytic,kennel1992fnn,fraser1986MI,10.1063/5.0137223}. 

Transformers and SSMs can both be viewed as delay embeddings, as they operate over a time history and are capable of inferring latent variables \cite{piantadosi2023chomsky}. However, the behavior of these sequence models depends on how they combine information across time. Transformer-based language models often sparsely combine inputs representing past states via a learned attention mechanism \cite{Li2023lazyneuron} while structured state space sequence models (SSMs) are designed to memorize as much of the past inputs as possible \cite{lmu,hippo,s4d}. Here, we apply the delay embedding perspective on transformers and SSMs to better understand the inductive bias of each architecture. In particular, we study the performance and dynamics of one-layer transformers and SSMs (the Linear Recurrent Unit \cite{orvieto2023LRU}) on a noisy, partially observed, chaotic dynamics prediction task. By forging the connection between delay embedding theory with sequence prediction in deep learning, we hope to establish a relationship that will mutually benefit research in both deep learning and dynamical systems theory.

\subsection{Contributions}
We characterize the embedding properties of 1-layer SSMs and transformers, showing that SSMs have a stronger inductive bias for delay embeddings, which leads to better attractor reconstructions and lower error on a chaotic prediction task. However, we also show that SSMs contain a large amount of redundancy, which excessively deforms the attractor and makes the model more sensitive to observational noise. While transformers do not have this inductive bias, we find that they are able to successfully learn a viable delay embedding with sufficient training.

%% file: methods.tex
\section{Methods}

We study one-layer sequence models, which consist of the following layers, in order: an encoding matrix, layer normalization, the sequence layer, layer normalization, and a three-layer MLP. We study one form of each class for simplicity: a GPT-style decoder-only transformer, and the linear-recurrent unit (LRU, \cite{orvieto2023LRU}). For the fairest comparison between architectures, we study transformers with positional embeddings applied \textit{only} within the softmax function of self-attention. Positional embeddings provide temporal information, breaking the permutation invariance of the transformer inputs to enable higher performance. Not including positional embeddings outside the softmax function, on the other hand, improves the quality of the manifold reconstructions. Embeddings were learnable. Equations describing each of the sequence layers can be found in the Appendix. 

\begin{figure}
\centering
\vspace{-10mm}
\includegraphics[width=\linewidth]{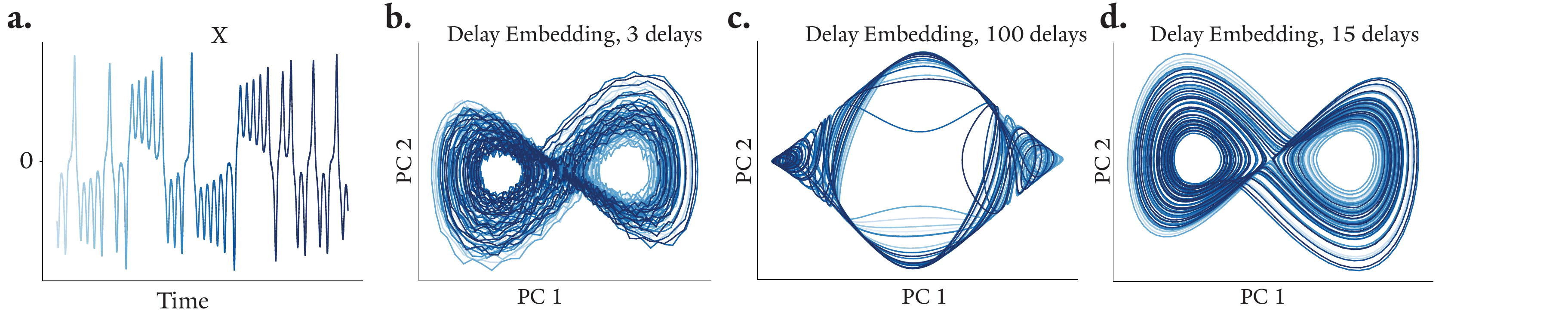}
\vspace{-8mm}
\caption{Delay Embeddings. \textbf{a.} Noisy data from the $x$ dimension of the Lorenz attractor, on which our models are trained. \textbf{b.} Visualization of the top two Principal Components of a delay embedding with too few delays. Here, noise is amplified and the attractor is distorted. \textbf{c.} As in \textbf{b}, with too many delays. Here the attractor is folded into too many dimensions, making the data harder to model. \textbf{d}. In the intermediate, the embedding both reduces noise and has a geometry that reflects the original space (visualized in Appendix \ref{lorenz}).}
\label{fig1}
\vspace{-8mm}
\end{figure}

Our models were trained on next-step prediction for a single observed variable of the 3-dimensional, chaotic Lorenz attractor (see Appendix \ref{lorenz}). We simulated 2000 trajectories of the system for 600 timesteps using the simulation timestep $\text{dt} = 0.01$ (Fig. \ref{fig1}a). Each trajectory had a unique initial condition. We removed the first 100 timesteps to eliminate transients. We added in i.i.d. Gaussian noise of zero mean and variance $0.1$ to the data at each timestep. We trained our models with Adam for 1000 epochs. For our analysis, we only studied models that reached a Mean Absolute Standardized Error (MASE) $< 1$. The MASE is the Absolute Error $|x_t - \hat{x}_t|$, normalized by the Persistence Baseline: $\hat{x}_t = x_{t-1}$.  MASE $\geq 1$ indicates that the model has captured no predictive information. At the time of submission, we trained approximately 25 networks of each architecture and dimensionality (10, 25, 50, and 100). We collected a similarly sized dataset for different values of observed noise (0.05 and 0.0), and display these results in Appendix Fig. \ref{perf_comparison}. 

\vspace{-4mm}
\subsection{Measuring Delay Embedding Quality} We quantified how well the sequence layer outputs operated as delay embeddings via three methods (futher detailed in Appendix \ref{metrics}): 
\paragraph{Decoding hidden variables}
We trained linear and nonlinear (MLP) decoders to predict the two unobserved dimensions of the 3-dimensional attractor, and measured the test $R^2$. 
\paragraph{Measuring Smoothness}
 Because a diffeomorphism is a smooth transformation, neighborhoods in one space should map onto neighborhoods in the other space. We measured this by identifying the fraction of overlap between the twenty nearest neighbors in the embedding and the true space. 


\paragraph{Measuring Unfolding}

Lastly, we measure how well the embedding lends itself to prediction, via the conditional variance of the future data given the embedding, $\sigma^2_\tau(x) = \text{Var}(x(t+\tau)|o_t)$ where $o(t)$ is the model hidden state. We provide implementation details in the appendix. We calculate the average of $\sigma^2_\tau$  over all data points, and average over $\tau$ from 1 to 10 steps in the future. 




%% file: results.tex
\section{Results}

\begin{figure}[h]
    \centering
    \vspace{-9mm}

    \begin{minipage}{0.48\textwidth}
        \vspace{-4mm}

        \centering
        \includegraphics[width=\linewidth]{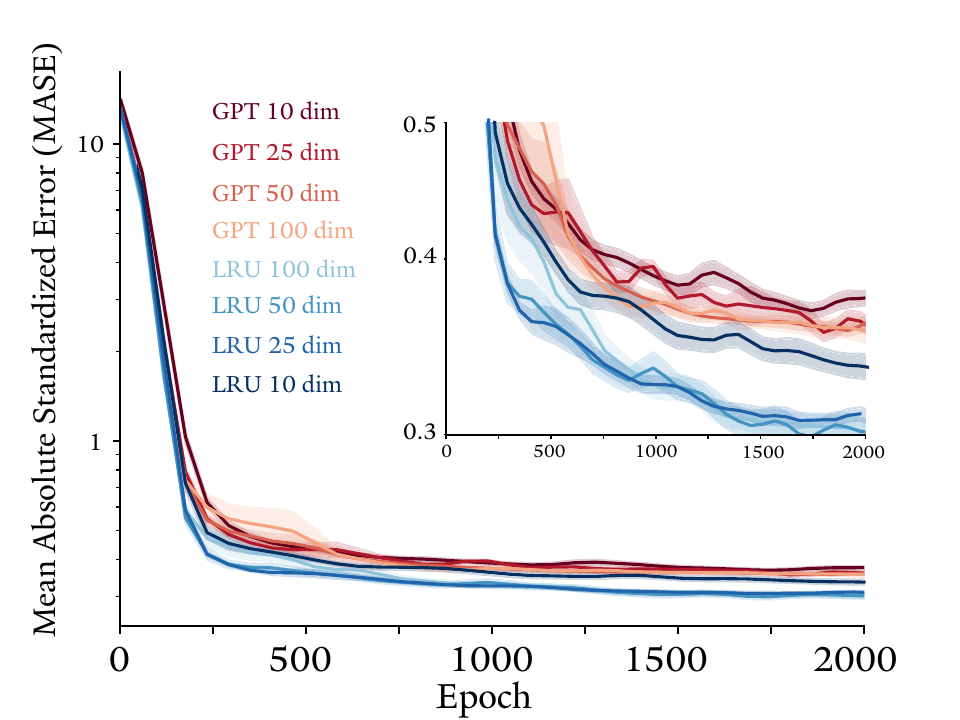} 
        \vspace{-8mm}
       \caption{Learning curves, inset zoomed in. Shading indicates standard error.}
        \label{fig2}
    \end{minipage}\hfill
    \begin{minipage}{0.49\textwidth}
        \centering
        \vspace{-4mm}
        \includegraphics[width=\linewidth]{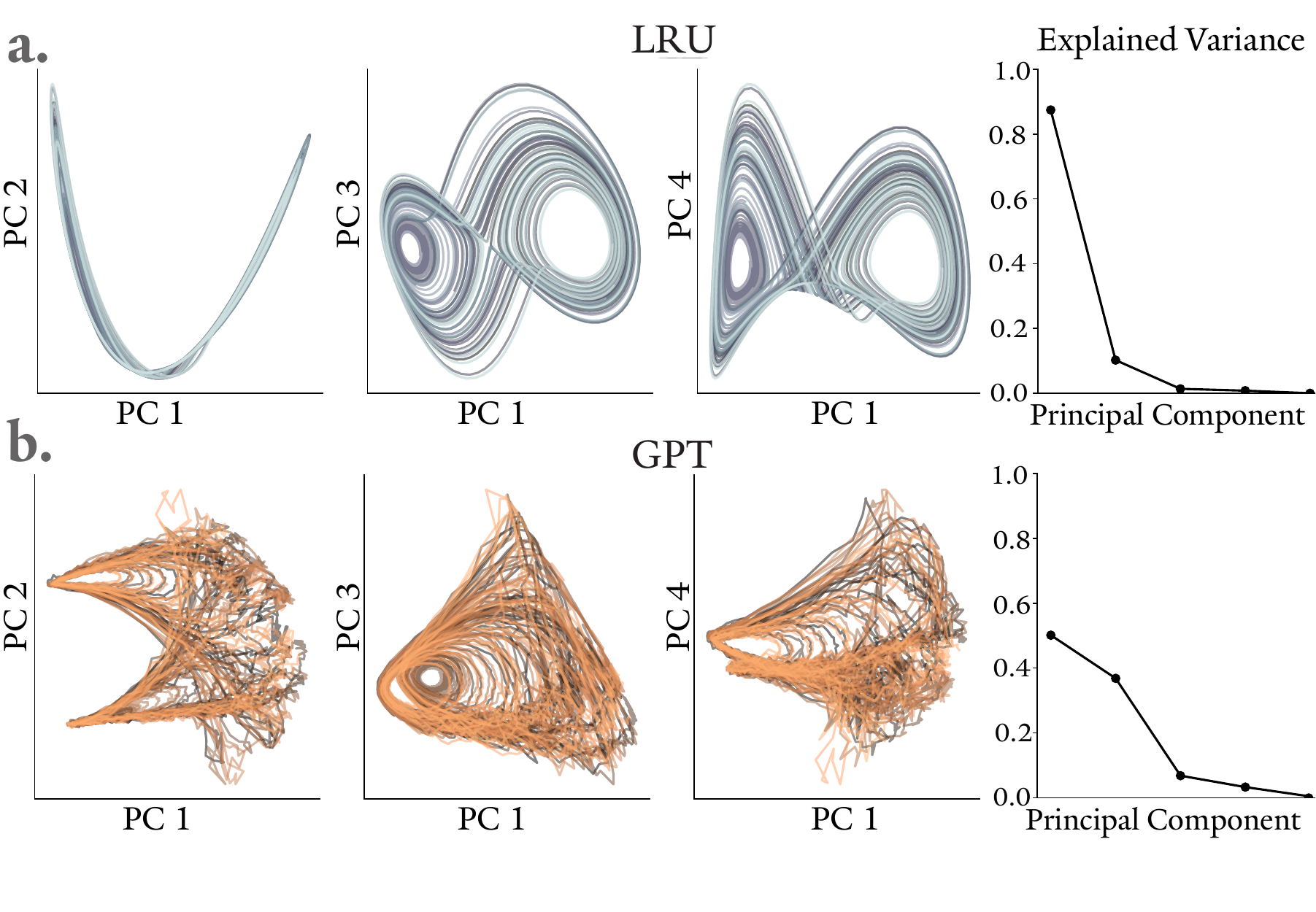}
        \vspace{-12mm}
        \caption{Sample top 4 Principal Components after training, of each sequence layer output, colored by trajectory.}

        \label{fig3}    
        \end{minipage}
\end{figure}





We began our analysis by inspecting the learning curves of each model, plotting the performance across training in Fig. \ref{fig2}. LRU models across all dimensionalities outperformed all GPT models in terms of MASE,  albeit by a small margin (final performance in Appendix Fig. \ref{perf_comparison}). 


To visually inspect the quality of embeddings, we plotted the top 4 Principal Components of two sample models that solve the noiseless task with MASE~$0.06$ in Fig. \ref{fig3} (LRU in \textbf{a}, GPT in \textbf{b}). Two observations are immediately evident: (1) The LRU embedding is more visually appealing, and (2) the butterfly shape emerged only in the LRU--this suggests that the LRU generated a more faithful delay embedding. However, in the first panel, PCs 1 and 2 suggest that this attractor is quite deformed. On the other hand, the transformer embedding is much less appealing. However, the two lobes of the attractor are identifiable in the first 2 PCs, and the dimensionality of the embedding (Participation Ratio of 2.47) is closer to the true attractor dimension (approximately 1.93) whereas the SSM embedding's dimension is 1.29. This larger dimensionality may lead to increased noise robustness of the transformers (see Appendix, Fig. \ref{noise_diff}).

\begin{figure}
    \vspace{-4mm}
    \begin{center}
        \includegraphics[width=\linewidth]{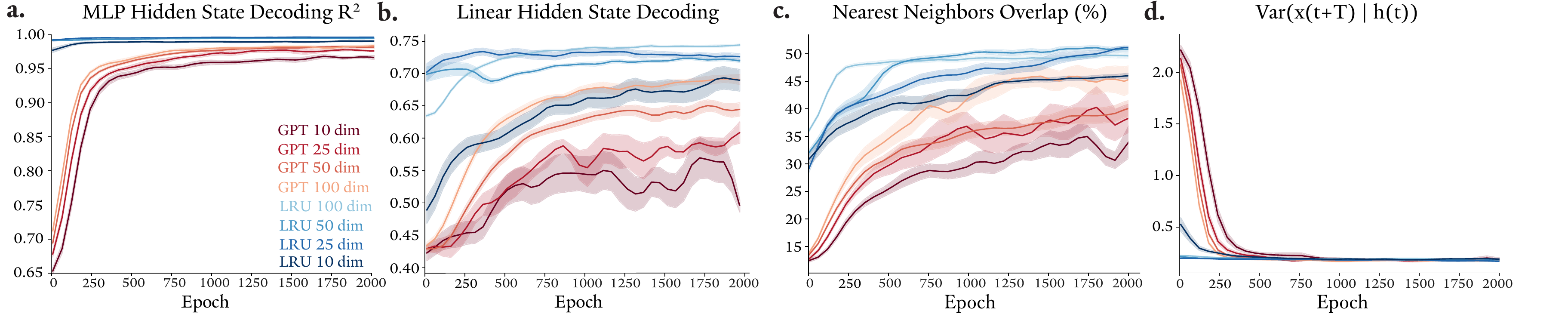}
    \end{center}
    \vspace{-8mm}
    \caption{Delay embedding metrics across training, colored by architecture and dimension. \textbf{a}. MLP decoding of unobserved variables, test $R^2$. \textbf{b}. Linear decoding test $R^2$. \textbf{c}. Neighbors overlap fraction between full dynamic state and embedding. \textbf{d}. Conditional variance of future data given the embedding, averaged over future time steps from 1 to 10. Shading indicates standard error.}
    \vspace{-8mm}
    \label{fig4}
\end{figure}

We applied each of the embedding metrics to the systems every 50 epochs during training, and plotted the results in Fig. \ref{fig4}, averaging over runs and separated by dimensionality and architecture. Across all metrics, we found that the LRU consistently started off with superior embedding quality. This indicates that the architecture has a powerful inductive bias for delay embedding. Importantly, each metric improved across training, demonstrating that the embedding can be optimized for better performance. The transformers, on the other hand, started off with much worse embedding metric performance, but gradually approached the embedding quality seen in the LRU models. We correlated each metric to prediction performance, finding a robust correlation between prediction quality and each of nonlinear decoding (correlation of 0.76), linear decoding (correlation of 0.56), and nearest neighbor overlap (correlation of 0.64, see Appendix \ref{performance}).

\vspace{-4mm}

%% file: discussion.tex
\section{Conclusion}

In this study, we demonstrated that the inductive bias of SSMs leads to improved embedding reconstruction, which was correlated with better performance on time series prediction. We found that SSMs have slightly higher performance on a dynamical systems prediction task, but were more sensitive to noise. Furthermore, for similar embedding dimensionality, the use of positional embeddings increases the parameter count of transformers relative to SSMs. This suggests that SSMs may be preferred in the low-data, low-compute regime. While our models and task were simplified, our study identifies a generic property of how time series are combined in each architecture, which is relevant for any application of these models to time series prediction.  In future work, we plan to study how transformers and selective SSMs \cite{gu2023mamba} select their delays.

%% file: appendix.tex
\section{}

\subsection{Lorenz attractor equations}\label{lorenz}

\begin{equation}
\frac{dx}{dt} = \sigma(y-x)
\end{equation}

\begin{equation}
\frac{dy}{dt} = x(\rho - z) - y
\end{equation}

\begin{equation}
\frac{dz}{dt} = xy - \beta z
\end{equation}

Where $\sigma = 10$, $\rho = 28$, $\beta = 8/3$. The fractal dimension of the attractor, measured using the Kaplan-Yorke dimension, is approximately 2.06. The dimension of the attractor, which we computed via participation ratio, is 1.986.

\begin{figure}[!h]
    \includegraphics[width=\linewidth]{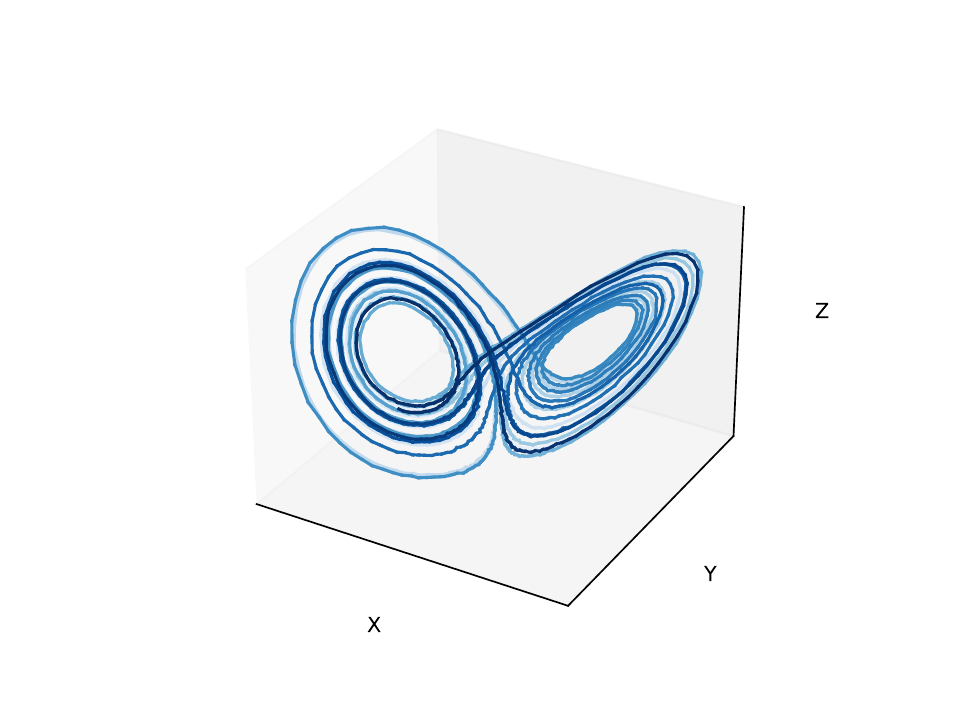}
    \caption{3-dimensional visualization of the Lorenz attractor. Simulated with noise, colored by time.}
\end{figure}

\subsection{Sequence Model Equations}

For a given sequence of inputs $U$, positional embeddings $P$, present time point $T$,  our self-attention layer is written as
\begin{equation}
    o_T = \text{SoftMax}[(U_{t \leq T} + P_{t \leq T})^TW^{qk}(u_t + p_t)]W^{ov}U_{t \leq T}
\end{equation}
Notably, taking positional embeddings out of the output (i.e., $W^{ov}(U_{t \leq T} + P_{t \leq T})$ does not negatively affect the performance, as the dynamics of the Lorenz attractor is time-invariant (it is an autonomous system). For convenience, we write the key and query matrices together as $W^{qk}$ and the output and value matrices together as $W^{ov}$.

The Linear Recurrent Unit layer \cite{orvieto2023LRU} is a discrete linear dynamical system:
\begin{equation}
    x_{t+1} = Ax_t + Bu_t
\end{equation}
\begin{equation}
    o_{t} = C\text{Re}(x_t) + Du_t
\end{equation}
In particular, $A$ is a diagonal matrix with complex eigenvalues initialized uniformly on a disk within the unit circle of the complex plane. This gives the model rotational dynamic properties, thereby allowing each input to be moved to a different subspace and be preserved. Because the input and output are real, the complex component of $x$ is discarded before a linear map to the output. 

\subsection{Embeddings before training}

\begin{figure}
    \centering
    \includegraphics[width=\linewidth]{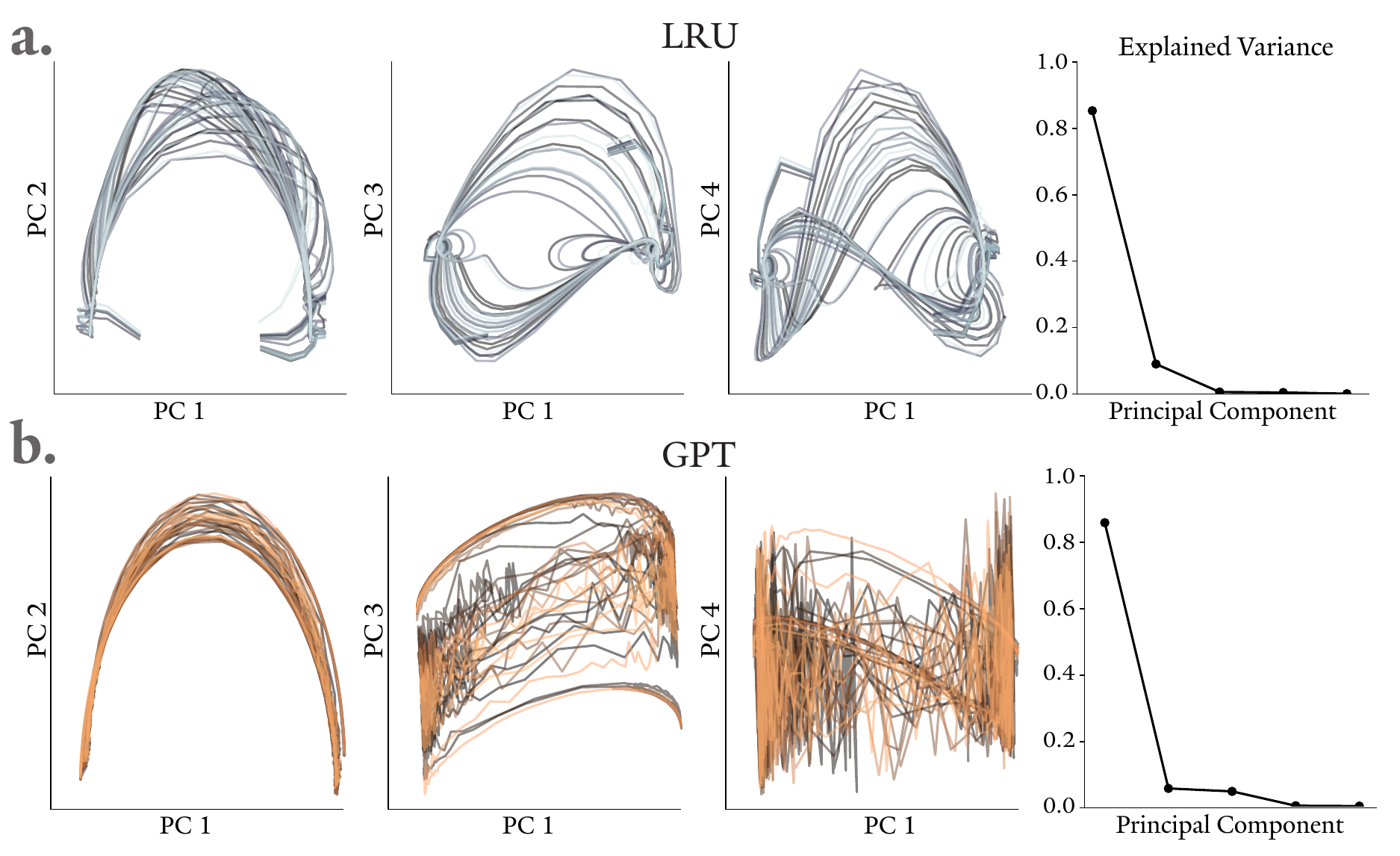}
    \caption{Untrained trajectories from each architecture, the same models as in Fig. \ref{fig3}.}
    \label{untrained_trajs}
\end{figure}

In Fig. \ref{untrained_trajs}, we visualize the embeddings of each sequence layer before training, when driven with a noiseless input. While neither looks much like the lorenz attractor, and are both quite low dimensional, it is evident that the LRU has a more similar appearance, with the two lobes evident in the 2nd plot. 

\subsection{Metrics}\label{metrics}

\subsubsection{Participation Ratio}

The Participation Ratio is a continuous measure of dimensionality, derived from Principal Components Analysis. While it is not equivalent to the fractal dimension of an attractor, which is typically measured via the Lyapunov exponents and is much more challenging to compute from data, it is useful to quantify dimensionality without setting arbitrary thresholds on the explained variance. Given the eigenvalues  $\lambda_i$ of the centered correlation matrix, the participation ratio is calculated as 

\begin{equation}
    p = \frac{(\sum_i \lambda_i)^2}{\sum_i (\lambda_i)^2}
\end{equation}

\subsubsection{Mean Absolute Standardized Error}
The MASE for an individual time point measures the performance of a predictive model, relative to the persistence baseline--the prediction one would make if they had no information. MASE is computed as follows:
\begin{equation}
\text{MASE}(x_t,\hat{x}_t) = \frac{\left | x_t - \hat{x_t} \right |}{\left | x_t - x_{t-1} \right |}
\end{equation}

\subsubsection{Convergent Cross Mapping}

Convergent Cross Mapping (CCM, \cite{sugihara2012detecting}) was developed as a measure of causality between dynamical systems, but here we chose to use it to measure continuity of the embedding, as it uses similar underlying principles. Given simultaneously recorded time series data from two dynamical systems ($x$ and $y$), CCM constructs a delay embedding of each, then uses the $k$-nearest neighbors (with $k$ equal to the embedding dimensionality) in one system to predict the value of the state in the other system. More specifically, given a particular time point $x(t)$, the k-nearest neighbors of the first system, $U(x)$, are identified. These points are then mapped to the embedding $y(t)$ yielding an equivalent set of neighbors, $U(y)$, (given the one-to-one mapping), and the prediction is made via their average: 
\begin{equation}
\hat{y}(t) = \frac{1}{k}\sum_{i \in U(y)}^k y_i
\end{equation}
We used the following code: https://github.com/nickc1/skccm.

\subsubsection{Neighbors Overlap}

We also implemented a stronger metric of continuity via the $k$-nearest neighbors, which we call the Neighbors Overlap. For each given point, mapped one-to-one via the embedding: $y = f(x)$, we compute the nearest neighbors of the true data space and the embedding separately: $U(x)$, $V(y)$. Then, we identify the time indices of each neigbor: $T_u, T_v$. The metric averages the fraction of overlap between these index sets across each point in the dataset:
\begin{equation}
    \text{Overlap}(X,Y) = \frac{1}{n}\sum_i^n \frac{\left | T_u(i) \cap T_v(i) \right |}{k}
\end{equation}
Where $\left | T_u(i) \right | = \left | T_v(i) \right | = k = 20$, $\left | X \right | = \left | Y \right | = n$. 

\subsubsection{Unfolding Metric}

We implemented the embedding complexity metric from \cite{uzal2011noise}. For noisy data, this metric characterizes the noise robustness of the embedding, and more generally calculates the complexity of the model required for next-step prediction. Here, we explain the motivation for the metric and detail the computational steps required to implement it. 

In Casdagli et al. 1991 \cite{casdagli1991reconstruction}, the authors suggest that predictive value is a good quantity to optimize for in an embedding. The authors define the predictive value of a reconstructed coordinate $y$ as the conditional probability density on the time-series values $T$ time-steps ahead:

$$p(x(t + T) | y(t))$$

\noindent where $x \in \mathbb{R}^D$ is the observed time-series and $y \in \mathbb{R}^d$ is the reconstructed coordinate. As noted by the authors, this quantity is independent of the predictive estimation procedure, as it captures what can be predicted about $x(t + T)$ from $y(t)$ with a perfect estimation procedure. The authors then go on to suggest that the variance of this distribution, given by

$$\text{Var}(x | y) = \int x^2 p(x|y) dx - \left ( \int x p(x|y) dx \right )^2$$

\noindent is a reasonable quantity to optimize for. Given that the ideal predictor is given by $\hat{x} = E(x | y)$, the above variance is the mean-square prediction error of the ideal predictor, and thus presents a lower bound on the mean-square prediction error of any predictor.

Building on this idea, Uzal et. al. \cite{uzal2011noise}  define the unfolding metric, which aims to estimate this variance. Given a time series dataset, the unfolding metric calculates two values for each time step based on its $k$-nearest neighbors. The first measures the variance of these points in the input space as time progresses. The latter is the volume of these points in the embedding space.

The conditional variance of future time steps, given the embedding, is approximated using the nearest neighbors: 
\begin{equation}
\text{Var}(x(T) | B_{\epsilon}(\tilde{x})) \approx E^2_k(T, \tilde{x}) \equiv \frac{1}{k+1} \sum_{\tilde{x}' \in U_k(\tilde{x})} [x' (T) - u_k(T, \tilde{x})]^2
\end{equation}
where $\tilde{x}$ is the delay embedded initial condition, $x'(T)$ is the future value of the true time series $x$ corresponding to initial delay embedded condition $\tilde{x}$, $B_{\epsilon}(\tilde{x})$ is a Gaussian ball with standard deviation $\epsilon$ around $\tilde{x}$, $U_k(\tilde{x})$ is the neighborhood of $k + 1$ points containing $\tilde{x}$ and its $k$ neighbors (and is an approximation of $B_{\epsilon}(\tilde{x})$, and the mean of the embedding is computed as:
\begin{equation}
u_k(T, \tilde{x}) = \frac{1}{k+1} \sum_{\tilde{x}' \in U_k(\tilde{x})} x'(T)
\end{equation}
Then, the overall conditional variance is computed by averaging $E^2_k$ over the first $p$ timesteps after the initial condition:

\begin{equation}
\sigma^2_k(\tilde{x}) = \frac{1}{p} \sum_{j=1}^p E_k^2(T_j, \tilde{x})
\end{equation}
where the $T_j$'s index the timesteps.



This is done for each individual data point, and we report the average across the attractor. We can also normalize this by the average volume of the whole attractor, which is equivalent to weighting the metric by the ergodic measure of the attractor. As suggested by \citet{uzal2011noise}, we used $k=3$ and $p = 10$.

\subsubsection{Dynamical Similarity Analysis}

DSA, developed in \cite{dsa}, is another method by which one could characterize the embedding nature of the sequence layer output. Briefly, the metric captures whether or not two systems are \textit{topologically conjugate}, the exact sort of similarity held between a system and its delay-embedded counterpart. However, because we mainly focused on the predictive capabilities of the embedding, both with respect to unobserved variables and the downstream task, we decided not to implement it here. 

\subsection{Relation of each metric to performance}\label{performance}

\begin{figure}
    \includegraphics[width=\linewidth]{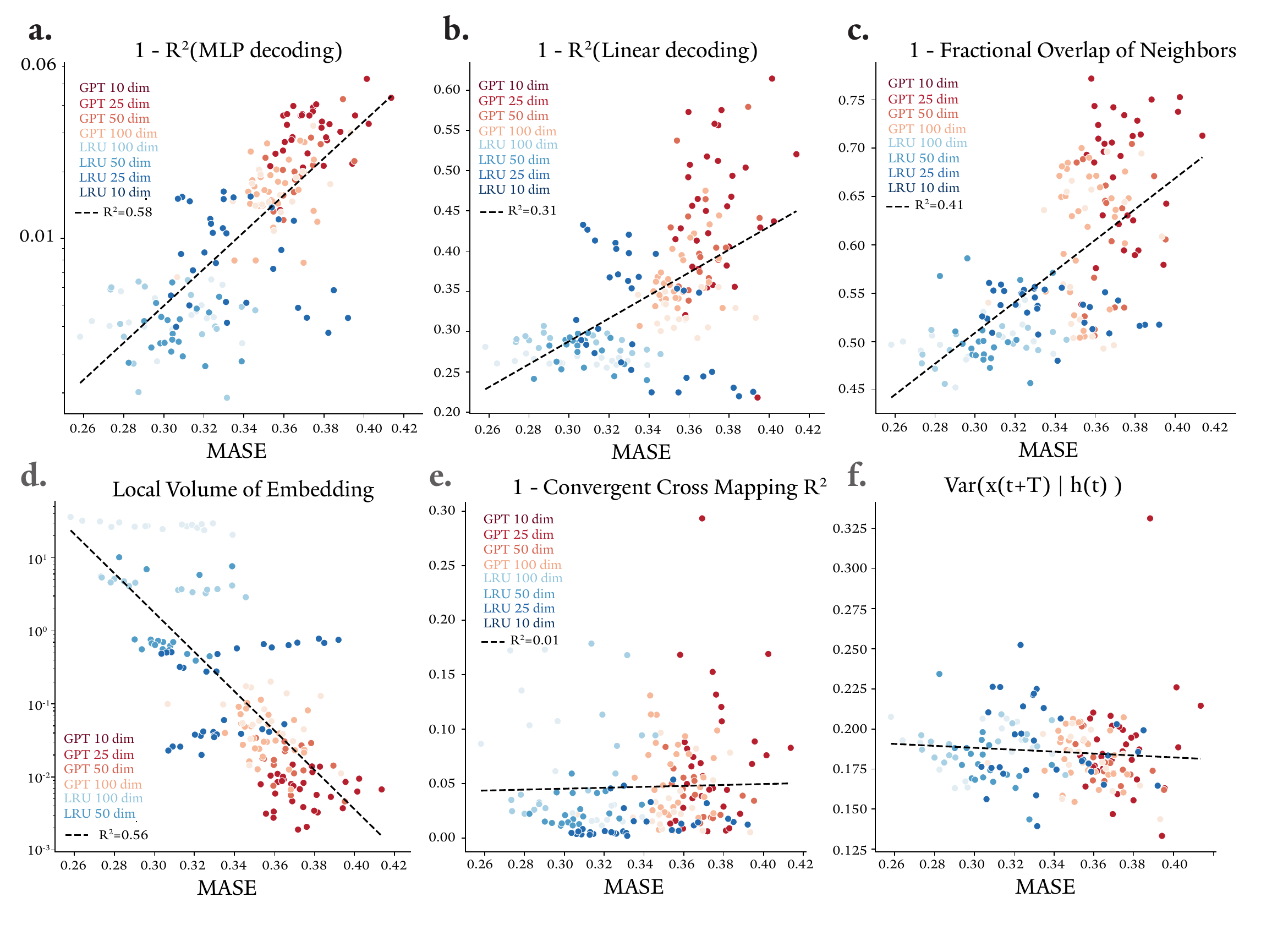}
    \caption{Relation of each embedding metric with performance. Each dot is an individual trained neural network, colored by architecture and dimensionality. $x$-axis on all plots is the validation Mean Absolute Standardized Error (MASE) after training. Dotted line indicates regression line of best fit, with $R^2$ coefficient listed in each legend \textbf{a.} MASE against nonlinear decoding accuracy $R^2$, plotted as $1-$ accuracy, due to logarithmic improvement. \textbf{b.} Likewise, MASE against linear decoding accuracy, demonstrating that better unfolding also improves performance. \textbf{c.} Relationship of MASE with fractional overlap of the 20 nearest neighbors in each embedding space (full data space versus output of the sequence layer). \textbf{d.} Relationship of the MASE with the local volume of the embedding, measured as the average distance of the 20 nearest neighbors in the embedding space from the present point. \textbf{e.} Relationship with the Convergent Cross Mapping Score, detailed in Appendix \ref{metrics}. \textbf{f.} Relationship with conditional variance, a measure of the unfolding complexity of the attractor, detailed in Appendix \label{metrics}. } 
    \label{metric_corr}
\end{figure}

Here, we demonstrate that in the noisy case, multiple embedding metrics are related to the model's predictive capacity. In Fig. \ref{metric_corr}, we scatter each metric for the noisy data, with noise variance $\sigma^2 = 0.1$. We observe a robust correlation for the nonlinear decoding accuracy, linear decoding accuracy, and neighbors overlap. This strengthens our hypothesis that a stronger embedding improves prediction performance, and provides insight into the superior predictive capabilities of the LRU models. 


As can be seen in Fig. \ref{metric_corr}f, we do not observe any correlation between performance and embedding complexity. This is likely because we utilize sufficiently wide MLPs for prediction, implicitly limiting the necessary complexity of the hidden embeddings. In future work, we will restrict the model expressivity and seek to identify a connection between embedding complexity and prediction performance.

\subsection{Change in performance due to noise}

\begin{figure}
        \centering
        \includegraphics[width=\linewidth]{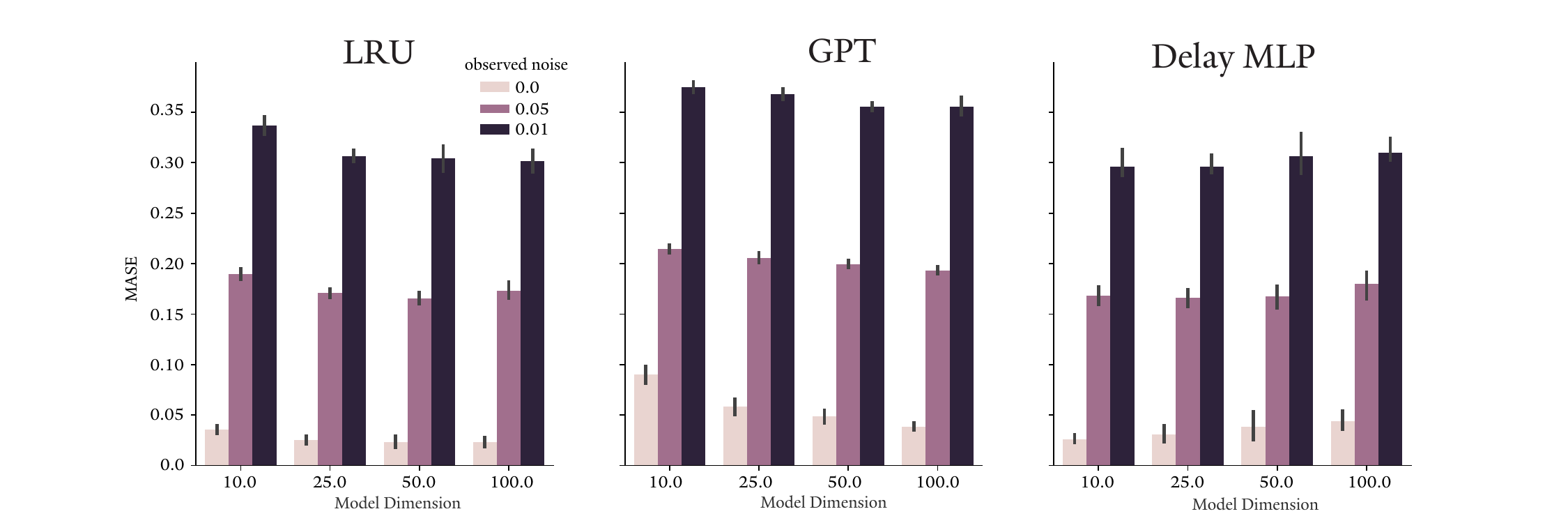} 
       \caption{Performance of each model across different dimensionalities and variance of observational noise. On the right, the baseline MLP over the delay embedding is displayed. Bars indicate standard error.}
        \label{perf_comparison}
\end{figure}

\begin{wrapfigure}{r}{0.45\textwidth}
    \includegraphics[width=0.5\textwidth]{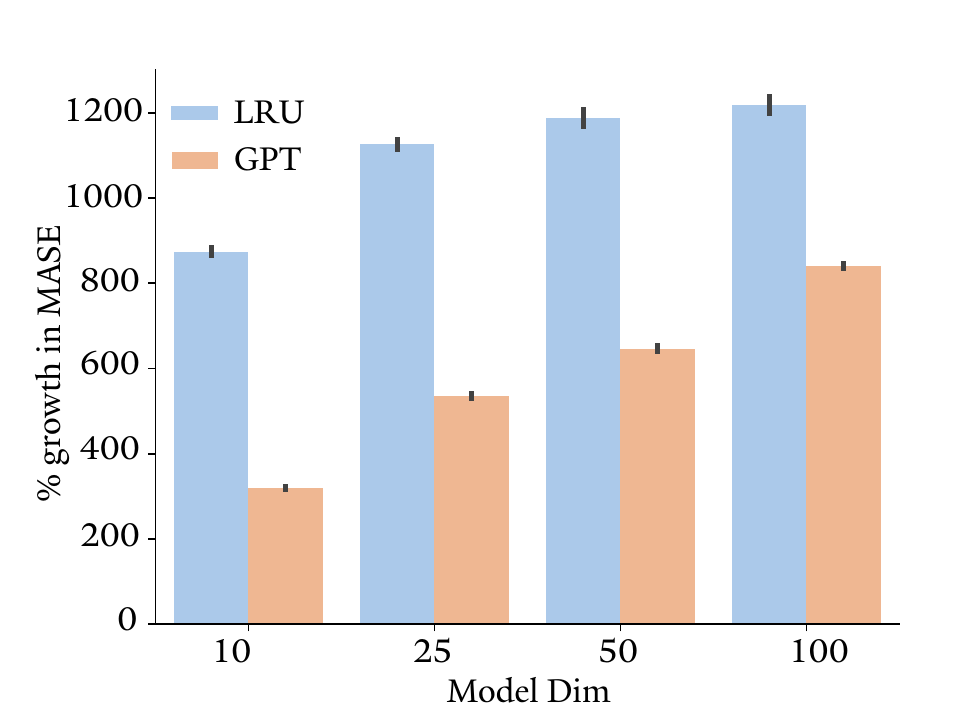}
    \caption{The percent growth in the MASE for the LRU and GPT when noise is increased to 0.1 from 0.0 (noise of 0.05 omitted for clarity). Bars indicate standard error.}
    \label{noise_diff}    
\end{wrapfigure}

To assess the robustness of each model architecture to observational noise applied i.i.d to each time point, we trained models with noise variances of $\sigma^2 \in \{0.0,0.05,0.1\}$. We display the performance of each model in Fig. \ref{perf_comparison}. We also train baseline models which we call 'Delay MLPs'--these are MLPs that operate over explicit delay embeddings. We simulated these with MLP widths ranging from 10 to 100, with a delay interval of 1 and a number of delays of sizes $\{10,25,50,100\}$. We find that the LRU models perform comparably to the Delay MLPs, and significantly better than GPTs across all noise levels. However, in Fig. \ref{noise_diff}, we measure the percent change in MASE when the noise is increased from $0.0$ to $0.1$, which shows that the LRU is more susceptible to noise than the transformer. This corroborates with results from Fig. \ref{fig3} which show that the LRU model is more highly folded and lower dimensional than the transformer. 

